\documentclass[journal]{IEEEtran}

\usepackage{graphicx}
\usepackage{amsmath,amssymb} % define this before the line numbering.
\usepackage{color}
\usepackage[numbers]{natbib}
\usepackage{listings}
\usepackage{multirow, subfigure}

\def\smallpics{1}

\begin{document}
	\title{LineNet: a Zoomable CNN for Crowdsourced High Definition Maps Modeling in Urban Environments}
	
	\author{
		Dun~Liang\IEEEauthorrefmark{1},
		Yuanchen~Guo\IEEEauthorrefmark{1},
		Shaokui~Zhang\IEEEauthorrefmark{1},
		Song-Hai~Zhang\IEEEauthorrefmark{1},
		Peter~Hall\IEEEauthorrefmark{2},
		Min~Zhang\IEEEauthorrefmark{3},
		Shimin~Hu\IEEEauthorrefmark{1}
		
		\IEEEauthorrefmark{1}Tsinghua~University
		\IEEEauthorrefmark{2}University~of~Bath
		\IEEEauthorrefmark{3}Harvard Medical School
		%	\thanks{TODO: thanks}
	}

	% The paper headers
	%\markboth{IEEE TRANSACTIONS ON IMAGE PROCESSING}%
	\markboth{}%
	{Dun \MakeLowercase{\textit{et al.}}: LineNet: a Zoomable CNN for Crowdsourced High Definition Maps Modeling in Urban Environments}
	\maketitle
	
	% As a general rule, do not put math, special symbols or citations
	% in the abstract or keywords.
	\begin{abstract}
		High Definition (HD) maps play an important role in modern traffic scenes. However, the development of HD maps coverage grows slowly because of the cost limitation. To efficiently model HD maps, we proposed a convolutional neural network with a novel prediction layer and a zoom module, called LineNet. It is designed for state-of-the-art lane detection in an unordered crowdsourced image dataset. And we introduced TTLane, a dataset for efficient lane detection in urban road modeling applications. Combining LineNet and TTLane, we proposed a pipeline to model HD maps with crowdsourced data for the first time. And the maps can be constructed precisely even with inaccurate crowdsourced data.
	\end{abstract}
	
	% Note that keywords are not normally used for peerreview papers.
	
	\IEEEpeerreviewmaketitle
	
	\section{Introduction}
	% The very first letter is a 2 line initial drop letter followed
	% by the rest of the first word in caps.
	% 
	% form to use if the first word consists of a single letter:
	% \IEEEPARstart{A}{demo} file is ....
	% 
	% form to use if you need the single drop letter followed by
	% normal text (unknown if ever used by the IEEE):
	% \IEEEPARstart{A}{}demo file is ....
	% 
	% Some journals put the first two words in caps:
	% \IEEEPARstart{T}{his demo} file is ....
	% 
	% Here we have the typical use of a "T" for an initial drop letter
	% and "HIS" in caps to complete the first word.
	
	% 但是高精度地图构建非常昂贵，现在主要有两种高精度地图的构建方式，一种是专业人员测绘，另一种是使用自己的车队和专业设备测绘（TODO:cite）。这两种方法都需要消耗大量的人力物力，如果要实现城市规模的高精度地图构建，这两种方法的成本无法接受，为了降低成本，我们使用了众包策略。我们开发了众包分发系统，让司机可以手机收集数据，上传到中心服务器，并获得一定收益。 众包策略除了降低成本以外，扩展和更新也很方便，但是同样也会给我们带来如下限制：

	With the development of navigation and driving applications, HD maps are in increasing demand. Yet modeling HD maps was very expensive. Currently, there were two main ways to build HD maps. One is engineering surveying and mapping. The other is using special fleets and professional equipment\cite{Bittel2017EstimatingHD, Mttyus2016HDMF} (.e.g, HD camera, deep sensor, radar, etc.). Both of these methods require a large amount of manpower and resources. However, to construct detailed HD maps on a city scale, the higher coverage area is required so that it is necessary to reduce costs. To enlarge the coverage area and reduce costs, we developed a crowdsourced distribution system. In our system, data acquired on mobile phones by drivers could also be integrated into the model and increase coverage area. In addition to reducing costs, the crowdsourced strategy is also very convenient to extend and update. But it suffers from following restrictions.
	
	\begin{itemize}
		% 受手机设备的限制，我们只能获取到相机图片和不准确的GPS信息。
		\item Due to the limitations of mobile devices, we can only obtain images and inaccurate GPS information.
		\item Images were captured by many drivers with unknown and different camera parameters.
		\item Images were sampled every 0.7 seconds by our crowdsourced app on the car due to the network bandwidth restriction. In an uneven speed, distance intervals of image sequences are different (usually 5-20 meters).
		\item Images of one road could be captured multiple times by different drivers.
	\end{itemize}
	
	Under these conditions, it is required to find a way to detect lanes accurately in a single image.
	Till now, there were only a few papers about road modeling \cite{Bittel2017EstimatingHD, Mttyus2016HDMF} and they did not consider crowdsourcing.
	So we proposed a road modeling system using crowdsourced images. First, we collect data in a crowdsourced way, then lane markings  are detected using neural networks, finally we model HD maps using  structure from motion (SfM) with the detection results.
	
	As for lane detection, all roads and all of the lane markings cannot be obtained simultaneously by current methods. There are currently two types of approaches. One is based on road surface segmentation mostly for ego-lane\cite{Teichmann2016MultiNetRJ,Chen2017RBNetAD,Oliveira2016EfficientDM}. However, these methods are  affected by occlusions and the types of lane boundaries are ignored. The other is based on lane markings \cite{Jung2013AnEL,
		berriel2017imavis,Beyeler2014VisionbasedRR,He2016AccurateAR,Lee2017VPGNetVP}. Lane markings contain more information, but they become narrower towards the vanishing points and finally converge, making it difficult to distinguish them. 
	%		in the scene.\textcolor{red}{Width of a lane marking is about five to ten pixels on the proximal side and one to two pixels on the remote side} \textcolor{blue}{I do not know what this means, please rephrase.}
	
	To detect all roads and classify all the individual lane markings at the same time, we need a method which can capture both the global view and details. As deep convolutional neural networks(CNN) have shown great capability in lane detection \cite{Lee2017VPGNetVP,He2016AccurateAR,Li2017DeepNN,Kim2017EndToEndEL}, we built a highly accurate CNN, named LineNet. To achieve high precision, a novel Line Prediction(LP) layer is included to locate lane markings directly instead of calculating from segmentation. In addition to the LP layer, the Zoom Module is added to recognize occlusion segments and gaps inside dashed lanes. With this module, the field of view (FoV) can be enlarged to arbitrary size without changing the network architecture. Combining these strategies together, LineNet can detect and classify lane markings across all roads. Details of LineNet will be discussed in Section~\ref{sec:linenet}.

	Existing lane detection datasets are not effective enough for HD maps modeling. Necessary road boundaries and supportive occlusion area for HD maps modeling are not counted in existing dataset, including KITTI Road \cite{Fritsch2013ITSC}, ELAS dataset \cite{berriel2017imavis}, Caltech Lanes Dataset \cite{Aly2008RealTD}, VPGNet Dataset \cite{Lee2017VPGNetVP},  tuSimple lane challenge \cite{tusimple} and CULane\cite{pan2018SCNN}. Therefore, we build a new dataset with more than 10,000 images. In each image, comprehensive annotations for all roads including road boundaries, occlusions, 6 lane types and gaps inside dashed lines are extracted (see Table~\ref{tab:dataset}). Details are in Section \ref{sec:dataset}.

	\setlength{\tabcolsep}{4pt}
	\begin{table*}
		\begin{center}
			\caption{Comparison of the KITTI road\cite{Fritsch2013ITSC}, ELAS\cite{berriel2017imavis}, VPGNet Dataset\cite{Lee2017VPGNetVP}, Caltech Lanes Dataset\cite{Aly2008RealTD}, tuSimple Lane Dataset \cite{tusimple}, CULane\cite{pan2018SCNN} and our TTLane Dataset.}
			\label{fig:compare_dataset}
			\begin{tabular}{lllllll}
				\hline\noalign{\smallskip}
				Datasets & Annotation type & Lane mark types & All roads  & Road boundaries & Occlusion segments & Total amount \\
				\noalign{\smallskip}
				\hline
				\noalign{\smallskip}
				KITTI road\cite{Fritsch2013ITSC} & segmentation & no & no  & no & no & 191\\
				ELAS \cite{berriel2017imavis} & line & yes & no & no & no & $\approx$ 15000\\
				Caltech Lanes Dataset\cite{Aly2008RealTD}  & line & yes & yes  & no & no & 1224 \\
				VPGNet Dataset\cite{Lee2017VPGNetVP} & line & yes & yes & no & no & $\approx$ 20000\\
				tuSimple Lane Dataset\cite{tusimple} & line & no & yes  & no & no & 6408 \\
				CULane\cite{pan2018SCNN} & line & no & no  & line & no & 133235 \\
				TTLane Dataset & line & yes & yes & yes & partial(3000/13200) & 13200 \\
				\hline
			\end{tabular}
		\end{center}
		\label{tab:dataset}
	\end{table*}
	\setlength{\tabcolsep}{1.4pt}
	
	To test our method for HD maps modeling, we ran experiments in two steps: lane detection (Section~\ref{sec:ld_for_hd}) and HD map modeling(Section~\ref{sec:hdmap_exp}). The HD map modeling pipeline is composed of OpenSfM\footnote{https://github.com/mapillary/OpenSfM}, LineNet, and some post processing. In the first step, LineNet was trained on different datasets and compared with the state-of-the-art techniques. It turned out that our approach outperformed other methods under various evaluation metrics. In the second step, our whole HD maps modeling pipeline was tested in a small-scale experiment. In this experiment, images and GPS information were collected with different transportations in three  conditions(straight, turning, and crossroads). The ground truth of these roads was also collected. And HD maps was successfully constructed with crowdsourced data for the first time. Compared with the ground truth, our crowdsourced method can reduce the error to 31.3 cm, which is desirable.
	
	%我们进行了小规模的实验，在这个实验中，我们邀请了三位骑手，分别使用了三种最常见的交通工具（汽车，自行车，电动自行车），对4条道路进行了建模。三位骑手覆盖了不同的路段，他们之间相互有重叠的部分。我们间隔5-20米采集了一些图片，配合不准确的GPS信息（误差约为5米）和LineNet检测结果，我们使用了SfM加上一定的后处理，完成了这4条道路的HD map构建，可以发现，SfM很好的连接了重叠部分，后处理过程得到了光滑可靠的车道线。我们同时对这4条道路进行了准确的测绘，对比准确测绘的数值，我们的众包方法可以把误差减少到10厘米左右，值得一提的是，尽管手机设备的GPS误差很大，该方法仍然可以将误差减少到很小。这是第一次利用众包数据的HD map modeling，我们取得了理想的效果。
	
	In summary, our main contributions are as follows.
	\begin{itemize}
		\item We proposed LineNet, a CNN model consisting of an innovative Line Prediction Layer and Zoom Module, and achieved state-of-the-art performance on two main subtasks of lane detection.
		\item We developed TTLane Dataset, the most effective dataset for lane detection in road modeling applications.
		\item We proposed a pipeline to model HD maps with crowdsourced data for the first time and achieved great precision even when the crowdsourced data is very inaccurate.
	\end{itemize}

	\section{Related works}

	\subsection{Datasets}
	The KITTI road \cite{Fritsch2013ITSC} dataset provides two types of annotation: segmentation of road and ego-lane. The road area is composed of all lanes where cars can drive. And the ego-lane area is the lane where the vehicle is currently driving. The ELAS dataset \cite{berriel2017imavis} also has ego-lane annotations, and Lane Mark Types(LMT) are annotated. More than 20 different scenes (in more than 15,000 frames) are provided in the ELAS dataset.
	
	The Caltech Lanes Dataset \cite{Aly2008RealTD} contains four video sequences taken at different times of the day in urban environments. It is divided into four sub-dataset and contains 1225 images in total. \citeauthor{Lee2017VPGNetVP} proposed VPGNet Dataset\cite{Lee2017VPGNetVP}, consists of approximately 20,000 images with 8 lane marking types and 9 road mark types under four scenarios, different weather, and lighting conditions. Unlike the first two datasets, all lane markings of VPGNet Dataset are annotated. Another dataset is tuSimple lane challenge \cite{tusimple}, that consists of 3626 training images and 2782 testing images taken on highway roads. CULane\cite{pan2018SCNN} currently is the biggest dataset about multiple lanes detection. But it is not designed for HD map modeling, as there is only annotations of the road currently being driven on and the annotations of road boundaries are missing.
	
	Our dataset offers more considerable details than any existing dataset: LMTs are provided, and all lanes in the image are annotated. Furthermore, LMTs are more concise: white solid; white dash; yellow solid; yellow dash; and double lines are all included. Gaps inside dashed lines are also annotated. The differences among datasets are summarized in Table~\ref{tab:dataset} and illustrated in Fig~\ref{fig:compare_dataset}.
	
	Our dataset is the only dataset that offers annotations of road boundaries and occlusion segments. Road boundaries are necessary because lots of HD maps applications need them, and occlusion segments are useful for improving LineNet(Section~\ref{exp_occl}). Annotations of road boundaries and separately labeled double lines make our TTLane Dataset a challenging large-scale dataset for lane detection in urban environments.
	
	\begin{figure}
		\centering
		\if\smallpics1
			\includegraphics[height=5.5cm]{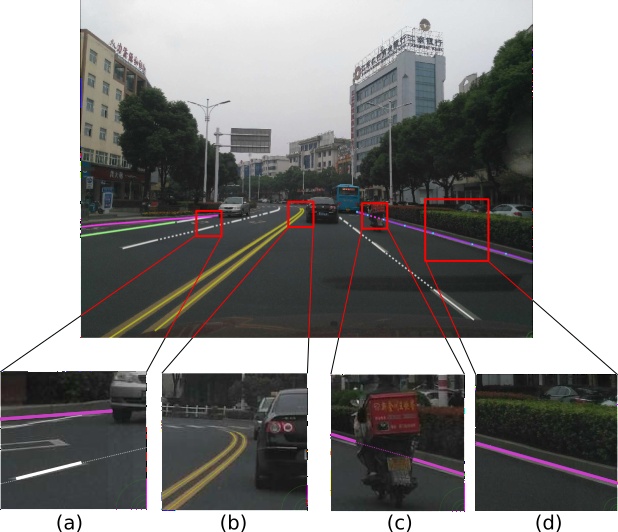}
		\else
			\includegraphics[height=5.5cm]{figs/anno.png}
		\fi
		\caption{This figure shows annotations of our TTLane Dataset: (a). Annotation of dash lanes. (b). Annotation of double lanes.
			(c). Annotation of occlusion segments. (d). Annotation of road boundaries.}
		\label{fig:annotation}
	\end{figure}
	
	\subsection{Lane detection with CNN}
	\label{subsec:algback}
	
	%	\textcolor{blue}{This section seems small. Since you compare with a general segmentation CNN this section could cite it. It should also  claim why general purpose segmentors are not suitable for the problem, and that you verify this claim experimentally. This will give you a second paragraph here.}
	
	Several papers about lane detection with CNN have been published in recent years  \cite{Lee2017VPGNetVP,He2016AccurateAR,Li2017DeepNN,Kim2017EndToEndEL}.
	\citeauthor{Lee2017VPGNetVP} \cite{Lee2017VPGNetVP} proposed a multi-task CNN to detect lanes and road marks simultaneously. They achieved the best F1-score in the Caltech Lane dataset \cite{Aly2008RealTD} because the  vanishing point was used. \citeauthor{He2016AccurateAR}\cite{He2016AccurateAR} combined features from the front view and Birds-Eye-View for lane detection. \citeauthor{Li2017DeepNN} \cite{Li2017DeepNN} extracted lane features from CNN, and fed them into a recurrent neural network, to utilize sequential contexts during driving. \citeauthor{Mttyus2016HDMF} \cite{Mttyus2016HDMF} used aerial images to improve the fine-grained segmentation result from ground view. Using the approach, they were able to recognize and model all roads. \citeauthor{xiaolong2017dilated}\cite{xiaolong2017dilated} proposed Dilated FPN with Feature Aggregation(DFFA) for drivable road detection. \citeauthor{Chen2017RBNetAD}\cite{Chen2017RBNetAD} proposed RBNet to detect both road and road boundaries simultaneously. DFFA\cite{xiaolong2017dilated} and RBNet\cite{Chen2017RBNetAD} achieved the best F1-score and precision respectively on the KITTI ego-lane segmentation task\cite{Fritsch2013ITSC}.

	It is important for lane detectors to have a large field of view (FoV). For example, understanding large gaps inside dashed lanes requires the FoV to be larger than the length of the gap.
	Many researchers \cite{Chen2017DeepLabSI,Wang2017ZoominNetDM,Xia2015ZoomBT,yularge,newell2016stacked} have shown that large FoV can increase performance; \citeauthor{Maggiori2016HighResolutionSL} \cite{Maggiori2016HighResolutionSL} introduced several techniques for high-resolution segmentation. \citeauthor{Chen2017DeepLabSI}\cite{Chen2017DeepLabSI} used atrous convolution \cite{Papandreou2015ModelingLA} and ResNet\cite{he2016deep} to achieve larger FoV, and \citeauthor{Wang2018non} \cite{Wang2018non} proposed non-local blocks to gather global information from the CNN. \citeauthor{newell2016stacked}\cite{newell2016stacked} proposed stacked hourglass networks, to gather a global context during the multi-level hourglass process. \citeauthor{yularge}\cite{yularge} proposed a global convolutional network to enlarge the FoV.
	\cite{Wang2017ZoominNetDM} proposed Zoom-in-Net that consists of three sub-networks. Low resolution images and high-resolution images are fed into two separate networks, and concatenated features are used for bounding box detection.
	
	All of the methods mentioned above have some shortcomings for HD maps modeling. DFFA\cite{xiaolong2017dilated} and RBNet\cite{Chen2017RBNetAD} predict the ego-lane segmentation, making LMTs lost. VPGNet\cite{Lee2017VPGNetVP} is hard to locate  non-parallel lanes or lanes that are too close. \cite{He2016AccurateAR, Li2017DeepNN} and \cite{Mttyus2016HDMF} require additional information such as perspective mapping. Our LineNet is designed to solve these issues.
	
	\setlength{\tabcolsep}{4pt}
	\begin{table*}
		\begin{center}
			\caption{Dataset Statistic. The meaning of these abbreviations is: WS(white solid), WD(white dash), RB(road boundaries), YS(yellow solid), YD(yellow dash)}
			\label{table:statistic}
			\begin{tabular}{llllllllll}
				\hline\noalign{\smallskip}
				width(px) & height(px) & images &WS & WD & RB & YS & YD & others  & total \\
				\noalign{\smallskip}
				\hline
				\noalign{\smallskip}
				$2058 \pm 163 $  & $1490 \pm 215$ & 13200 &24831 & 14136 & 22598 & 8945 & 2106 & 2613 & 75231\\
				\hline
			\end{tabular}
		\end{center}
	\end{table*}
	\setlength{\tabcolsep}{1.4pt}
	
	\subsection{HD Maps Modeling}
	There were only a few papers about HD Maps modeling\cite{Mttyus2016HDMF, Zang2017LaneBE, Bittel2017EstimatingHD}. Most of them have either\cite{Zang2017LaneBE,Bittel2017EstimatingHD} or both\cite{Mttyus2016HDMF} the following two characteristics: using aerial images\cite{Zang2017LaneBE} and with professional equipments\cite{Bittel2017EstimatingHD}. Using aerial images to model HD Maps has a natural advantage, which is that the geographic location and topology of the maps are easy to construct. But it also has shortcomings: the road may be covered, or the resolution of aerial images is not enough for HD Maps modeling. With professional equipment(mount on the car), HD maps can be modeled precisely, but the costs are too high to detailedly model HD maps on a city scale. Street view is a related topic of HD maps, most companies (.e.g Google, HERE, Tencent, etc.) collect data with own fleets, but coverage is not high enough especially in rural and remote areas, the street view coverage of Tencent maps grows slowly in recent years, because the cost would be unacceptable if we want to enlarge the coverage.
	To detailedly model HD maps on a city scale, we choose the crowdsourced strategy. It can overcome the two shortcomings mentioned above without losing accuracy. We have developed a crowdsourced distribution system that allows drivers to collect data on mobile phones, and the HD maps can be updated without delay. It is worth mentioning that we can use those methods together, and our crowdsourced method can focus on regions where own fleets are hard to reach.
	\section{TTLane Dataset}
	\label{sec:dataset}
	\subsection{Data Collection}
	
	Our dataset is crowdsourced. It was collected from different drivers(about 200), each with a phone camera mounted on the drivers' dashboard. Because the dataset was obtained from different drivers', phones, orientations, resolutions, exposures, focal lengths and so on, any algorithm should be robust to such variations. All images were collected in urban environments.
	
	Most exist datasets are continuous video sequences because the temporal information can improve lane detection results. However, our dataset is different, TTLane has fewer overlapping parts, most of them are unordered. There are two reasons for this. First, the diversity of our dataset can be increased, the data can be collected from all over the world. Second, crowdsourcing suffers from the bandwidth of the mobile network, dense video sequences require more network traffic.

	%asked to install a Tencent crowd-source  push the phone in front of the car, the APP will record the image and GPS and send the data back to the server. Our pictures are taken in urban area of Beijing, China. the phone are placed in the car, so wipers and stains on windows can interfere with the view sometimes. Because the dataset is collected from different driver and phone, orientation, resolution, exposure, focal length are different, it requires the algorithm should be robust.
	
	% we randomly sampled 1200 images from the entire dataset.
	% 1000 image for training set and 200 image for testing set.
	
	\subsection{Annotation}
	
	We manually annotated the center points of lane markings, Continuous lane markings are fitted and plotted with a Bezier curve. Each lane marking has a type. There are 6 types in total (white solid, white dash, yellow solid, yellow dash, road boundary, other). If there are double white solid lines in the road center, the annotator must draw 2 parallel white solid lines, which is shown in Fig~\ref{fig:annotation}. This annotation method can cope with complex double lane markings, such as left solid and right dashed lanes.
	
	%The reason why we chose to annotate double lane marking separately is, LMTs are much complicated in urban environments, dash-solid lane marking appeared in Beijing, those lane markings has special meaning. In order to handle those scenes, we decide to annotated separately for rich representation.
	
	%	\textcolor{blue}{Table 1 says you do not annotate the occluded area - this text says you do annotate the occluded area. We should distinguish between occlusions of the road by (say) other cars, and gaps inside dashes in dashed lines. Please consider this issue at check the entire paper to make sure the text is fully consistent throughout.}\textcolor{red}{\em(that was a mistake in Table 1, we did annotate the occluded area)}
	
	Annotations of occlusion segments and gaps inside dashed lanes are among other features of our dataset. \citeauthor{Liang2016ReversibleRI} \cite{Liang2016ReversibleRI} show that analysis of the occluded region can improve image segmentation performance, see more details in Section \ref{exp_occl}. So we annotated the occlusion segments with a dotted, continual lane. The same is true for gaps in dashed lanes. Note that we do not distinguish among occlusions and gaps -- and they were annotated in the same format. As is shown in Fig~\ref{fig:annotation}.
	\begin{figure}
		\centering
		\includegraphics[height=4.5cm]{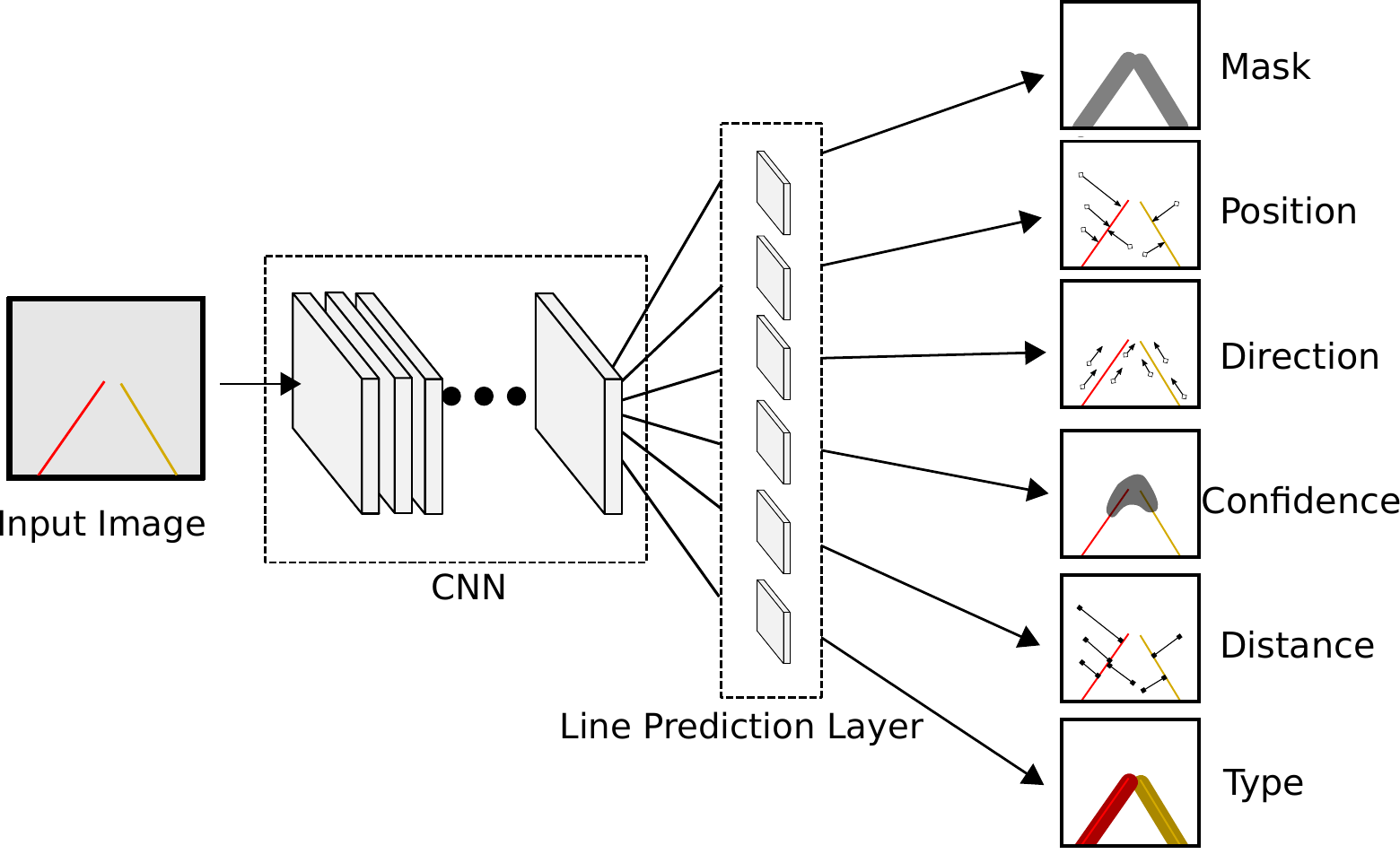}
		\caption{An illustration of the LP layer. LineNet will become confident when two lane markings are sufficiently far from each other.}
		\label{fig:lplayer}
	\end{figure}

	\subsection{Dataset Statistics}
	
	Our dataset consists of 13200 images, 3000 of which were annotated with occlusion information and the rest of the 10200 images were usually annotated. Isolation belts, roadblocks and other things that may affect the road conditions were annotated as ``other''.  Regarding comprehensive annotation, our dataset is the most detailed and complicated dataset for road modeling in urban environments.
	
	\section{LineNet}
	\label{sec:linenet}
	
	\subsection{Line Prediction Layer}
	\label{sec:lpl}

	Current lane detection investigations based on CNN were restricted to segmentation methods, which is not intuitive for line detection, causing inaccuracy. It is insufficient for real-life demands. Thus we propose a novel approach, that aims to provide new ideas for lane detection tasks.
	
	We use a pre-trained ResNet model with dilated convolution\cite{Chen2017DeepLabSI} as the feature extractor. a dilated convolution strategy helps to increase receptive field, which is essential when detecting dashed lanes. To make the model suitable for the lane detection, we make two non-trivial extensions:
	\begin{itemize}
		\item An additional layer designed for lane positioning and classification, called the Line Prediction(LP) layer.
		\item A zoom module that automatically focuses on areas of interest, i.e., converging  and intersection areas of the lane markings.
	\end{itemize}
	\begin{figure*}
		\centering
		\if\smallpics1
			\includegraphics[height=4cm]{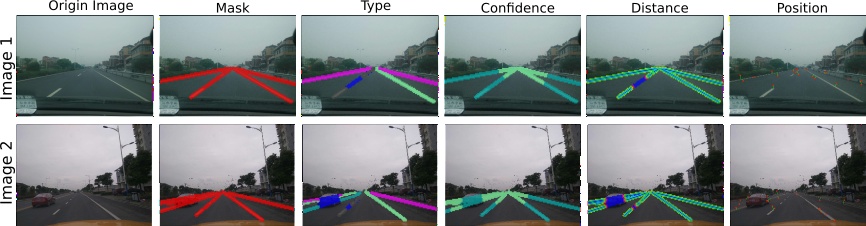}
		\else
			\includegraphics[height=4cm]{figs/lplf.png}
		\fi
		\caption{This figure shows different branches' outputs of the LP layer, with two samples. The first column shows the  original images. The second column shows the mask of the lanes. The background image in this row is much more blurred than the first column because this is what LineNet sees (i.e., the zoom level is 0.5 so we down-sample the input images). In the third column, different colors represent different lane marking types. Note the blue color represents the occlusion or gaps. In the fourth column, the light green color represents the unconfined area, and the cyan color represents the confident area where predictions are reliable. The fifth column represents the distance map from the anchor point to the center line. In the sixth column, we randomly sample some anchor points (blue points), and the green lines show the outputs of the position branch.
		}
		\label{fig:lplf}
	\end{figure*}
	
	The Line Prediction (LP) layer is designed for accurate lane positioning and classification, inspired by \citeauthor{Zhu2016TrafficSignDA}\cite{Zhu2016TrafficSignDA}'s three branch prediction procedure. There are six branches in our LP layer: line mask; line type; line position; line direction; line confidence; and line distance, as is shown in Fig~\ref{fig:lplayer}.
	
	A line mask is a stroke we draw with a fixed width (32 pixels in our experiments). The line type indicates one of the six-lane marking types. Line position predicts the vector from an anchor point to the closest point in the line. Supervising on the line position can produce much more accurate results than using a mask. Fig~\ref{fig:lplf} illustrates this by showing that the line position branch can predict lanes in sub-pixel level. Line distance is the distance from the anchor point to the closest point in the line. Line direction predicts the orientation of the lane. Line confidence predicts the confidence ratio, i.e., whether the network can see the lane clear enough, which is defined between 0 (if two lines are closer than 46 pixels) and 1 (otherwise). When we zoom into two adjacent lines, they will gradually separates, result in change of the confidence ratio. We will discuss how do the six branches work together to predict a whole lane in Section \ref{sec:post}. Figure~\ref{fig:lplf} shows the supervised value in the LP layer with two examples.
	
	\subsection{Zoom Module}
	\begin{figure*}
		\centering
		\includegraphics[height=5.0cm]{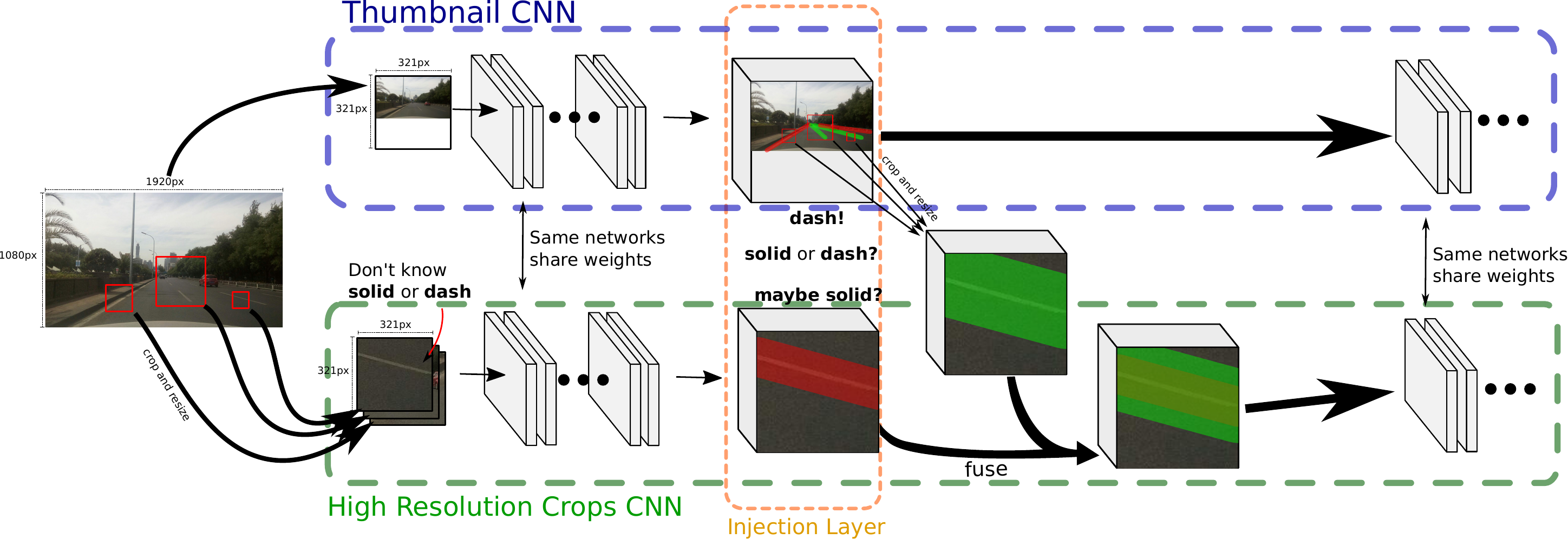}
		\caption{The Zoom Module fuses data from the thumbnail CNN
			and the high-resolution cropped CNN together, and detect a dashed lane in the
			high-resolution stream. The thumbnail CNN can recognize the dashed lane because it has larger FoV (field of view), while the high-resolution CNN accurately locates the lane markings. }
		%		for high-resolution Crops CNN that it can know the lane marking is dash or solid before injection layer. After injection layer, the "dash feature"(green) from Thumbnail CNN's injection layer has fused and covered with "solid feature"(red) from high-resolution Crops CNN. This operation gives the ability for the high-resolution Crops CNN to recogize the dash lane.}
		\label{fig:zoom}
	\end{figure*}

	%	\textcolor{blue}{The above might be better moved to related work (subject to appropriate editing) it will make the related work section larger and leave this section clear for your explanation. Be sure to say what the differences are between you zoom and the zoom of others, or at least explain why you wrote your own zoom.}
	%	\textcolor{red}{
	
	The Zoom Module is the second feature of LineNet. With this module, LineNet can  alter the FoV to an arbitrarily size without changing network structure. Our Zoom module is inspired by Zoom-in-Net \cite{Wang2017ZoominNetDM} and non-local block \cite{Wang2018non}. It splits the data flow through the CNN into two streams:  (i) a thumbnail CNN; and (ii) a high-resolution cropped CNN. Fig~\ref{fig:zoom} illustrates the architecture of the Zoom Module. Fig~\ref{fig:zoom} shows that the FoV depends on the size of the crops and thumbnails. A fixed network structure has lots of benefits, including effective use of pre-trained weights after Zoom Module is mounted, and fixed computational complexity.
	
	The thumbnail CNN provides a global context for the features that the high-resolution CNN ``sees'' in detail. These two CNNs share weights and compute in parallel, up to a point where information from the ``thumbnail'' CNN is shared with the high-resolution CNN. Sharing is achieved via a layer we call the ``injection layer'', that fuses features from both CNNs, and places the result in the data stream of the high-resolution CNN. 
	
	%This operation makes the high-resolution crops CNN see the global features from thumbnail CNN, without losing it's own clarity.
	
\begin{figure}
		\centering
		
		\if\smallpics1
		\includegraphics[height=7.5cm]{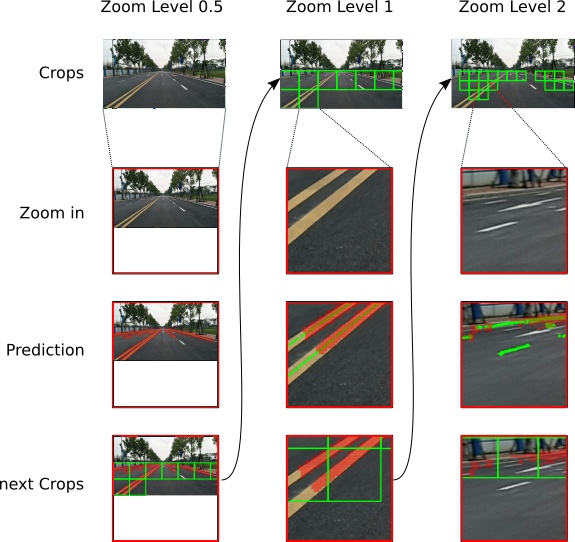}
		\else
		\includegraphics[height=7.5cm]{figs/zoomp.png}
		\fi
		\caption{This figure illustrates the zooming process. Three columns represent three different zoom levels (more zoom levels can be added if necessary). The first row shows the high-resolution crops are fed into LineNet. The second row shows what LineNet sees in one crop. And the third row shows the prediction of LineNet in this crop. The red area represents the uncertain area. The fourth row represents crops that are generated for the next zoom level.}
		\label{fig:zoomp}
	\end{figure}

	%Zoom in blindly is very time consuming, in order to improve the complexity,
	The Zoom Module and LP layer work together to boost performance. The zoom module allows the network to zoom into areas where the LP layer is not confident enough. In practice, the zooming procedure is used multiple times, with the zoom ratio gradually multiplied from 0.5 to 16. Fig~\ref{fig:zoomp} shows the four stages of the zooming process and how the LP layer and the zoom module interact with each other through the output of the line confidence branch. Overall, when the certainty of the lane rises, the spatial location of the lane becomes more accurate.
	
	\subsection{Post Processing}
	\label{sec:post}
	
	The results that LineNet detected were still discrete points. To achieve nice and smooth lines, points were clustered together and fitted into lines. The clustering algorithm named DBSCAN\cite{Ester1996ADA} was used with our hierarchical distance(HDis). The line position from the LP layer was collected and combined with a zoom level. The combination is denoted as a tuple $a = (x,y,z)$, where $(x,y)$ is the image coordinate from line position outputs, and $z$ is the stage's zoom ratio used to predict the line position. We define the distance of two points, $a$ and $b$, in the equation~\ref{equ:dis}. This distance was applied in DBSCAN clustering. And it allows DBSCAN to manipulate the appropriate distance from different zoom ratios. 
	
	\begin{equation} \label{equ:dis}
	HDis(a,b) = max(a.z,b.z)\sqrt{(a.x-b.x)^2 + (a.y-b.y)^2 }
	\end{equation}

\begin{figure}
		\centering
		
		\if\smallpics1
		\includegraphics[height=4.5cm]{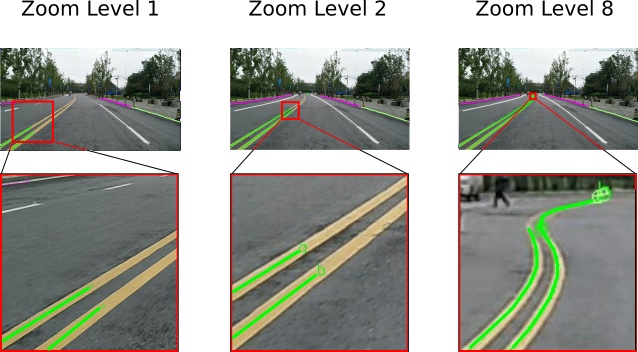}
		\else
		\includegraphics[height=4.5cm]{figs/zc.png}
		\fi
		\caption{This figure illustrates that line points are gradually clustered together from near to far. Different colors represent different line types.}
		\label{fig:post}
	\end{figure}

	After DBSCAN\cite{Ester1996ADA} clustering, line points of each cluster were fitted into a polynomial $p(x) = p_0 + p_1 x + p_2 x^2 + p_3 x^3 $, then smooth and reliable lines were achieved the and ready for evaluation. Fig~\ref{fig:post} shows that the lane results are gradually clustered together during the zooming process.

	\section{Experiments}
	\label{sec:exp}
	
	We divide our experiments into three subsections: 1. lane detection for HD maps modeling, that evaluates the effectiveness of our LineNet; 2. HD maps modeling, that analyzes the accuracy of our modeling pipeline; 3. Ablation studies.
	
	\subsection{Lane Detection for HD Maps Modeling}
	\label{sec:ld_for_hd}
	There are  two common subtasks in the lane detection task, which are:  lane positioning and comprehensive lane detection. To prove the effectiveness of our method, we apply our model to the two subtasks and also compare our method with state-of-the-art approaches. The experiment of HD map modeling is described in Section~\ref{sec:hdmap_exp}.
	
	\subsubsection{Implementation Details}
	Our implementation is based on Tensorflow. We set the learning rate as the base learning rate multiplying $(1 - \frac{iter}{iter_{max}})^{power}$; base learning rate is $2e^{-4}$, and the power is $0.9$. We use $0.9$ for momentum and $5e^{-4}$ weight decay. The number of training iterations varies due to different dataset sizes, but $80000$ iterations are sufficient for the three datasets with which we evaluate.
	
	For the zoom module, we use the 66th layer (4b13) of the Dilated Resnet architecture \cite{Chen2017DeepLabSI} as the injection layer, Experiments showed that average fusion works well among various fusion methods.

	\subsubsection{LineNet for Lane Positioning}
	In the lane positioning subtask, our goal is to predict all the lane markings in the image without considering their types. We use Caltech Lanes Dataset and our TTLane Dataset for this subtask. Caltech Lanes Dataset \cite{Aly2008RealTD} contains 1225 images taken at different times of a day in urban environments. It is divided into four sub-datasets, cordova1, cordova2, washington1, and washington2, that contain 250, 406, 337 and 232 images respectively. We used the same training and testing set as in \cite{Lee2017VPGNetVP}. Our TTLane Dataset has up to 10K high-resolution finely annotated images taken in urban environments. We use 13000 images for training and validation and 200 images for testing and evaluation. Both datasets are annotated in a line-based way.
	
	We adopted the method mentioned in \cite{Lee2017VPGNetVP}, {\em i.e.}, meaning we drew ground truth lines and predicted lines with a thickness of 40 pixels and evaluate the performance in a pixel-based way using F1-score measure.
	
	We also adopted the method mentioned in \cite{pan2018SCNN}, they draw lane markings with widths equal to 30 pixels and calculate the IoU between ground truth and predictions, we consider 0.5 as the threshold with the F1-score measure.
	
	\setlength{\tabcolsep}{4pt}
	\begin{table*}
		\begin{center}
			\caption{Comparison of lane positioning.}
			\label{table:caltech}
			\begin{tabular}{l|c|l|c|l|ccc}
				\hline
				\multirow{2}{4em}{Method} & \multicolumn{1}{|c|}{CULane} &
				\multirow{2}{4em}{Method} & \multicolumn{1}{|c|}{Caltech Lanes} & \multirow{2}{4em}{Method} & \multicolumn{3}{|c}{TTLane} \\
				\cline{2-2}\cline{4-4}\cline{6-8}
				&\multicolumn{1}{|c|}{F1-score} & & \multicolumn{1}{|c|}{F1-score} & & \multicolumn{1}{|c}{F1-score} & \multicolumn{1}{c}{Precision} & \multicolumn{1}{c}{Recall} \\
				\hline
				Ours & \textbf{0.731} & Ours & \textbf{0.955} & Ours & \textbf{0.832} & \textbf{0.848} & \textbf{0.816} \\
				SCNN\cite{pan2018SCNN} & 0.713 & VPGNet\cite{Lee2017VPGNetVP} & 0.866 &Mask R-CNN\cite{He2017MaskR} & 0.708 & 0.764 & 0.660 \\
				& & Caltech\cite{Aly2008RealTD} & 0.723 & MLD-CRF\cite{Hur2013MultilaneDI} & 0.412 & 0.556 & 0.327 \\
				& & & & SCNN\cite{pan2018SCNN} & 0.790 & 0.794 & 0.787 \\
				\cline{1-8}
			\end{tabular}
		\end{center}
	\end{table*}
	\setlength{\tabcolsep}{1.4pt}
	
	We compare our method with VPGNet\cite{Lee2017VPGNetVP}, Mask R-CNN\cite{He2017MaskR}, SCNN\cite{pan2018SCNN} and MLD-CRF\cite{Hur2013MultilaneDI}. VPGNet was the state-of-art method for comprehensive lane detection. Mask R-CNN is one of the best-performed object detection methods. MLD-CRF uses a conditional random field to do line positioning, and it is considered one of the best-performed method that do not need training. And SCNN\cite{pan2018SCNN} won the 1st price in TuSimple lane detection challenge 2017\cite{tusimple}. Results are shown in Table~\ref{table:caltech}. Our method achieved the best performance, and the result even exceeded the prior best performance by a substantial margin. This achievement reflects that our method works very well in line positioning.
	
	\subsubsection{LineNet for Comprehensive Lane Detection}
	\label{sec:cld}
	In this subtask, we need to predict both lanes marking positions and their types. It is  the most difficult subtask of all two subtasks we did experiments on, especially in complex scenes where there are moving vehicles and other actions.
	
	Our TTLane Dataset is a more challenging dataset for up to 6 different line types. We used 13000 images for training and validation, and 200 images for testing and evaluation, as is in the line positioning subtask.
	Unlike the metrics we mentioned above, here we evaluate area and distance between lines, not masks. We adopted a method that is similar to the one in \cite{Aly2008RealTD}. Consider the ground truth line as \texttt{gtLine} and the prediction result  as \texttt{pdLine}. The prediction result is considered correct when both of the following two conditions are met:
	
	\begin{itemize}
		\item The distance between the endpoints of two lines is smaller than a certain threshold (annotated as \texttt{disThresh}).
		\item The area enclosed by two lines is smaller than a certain threshold (annotated as \texttt{areaThresh}).
	\end{itemize}
	
\begin{figure}
		\centering
		\includegraphics[height=2.5cm]{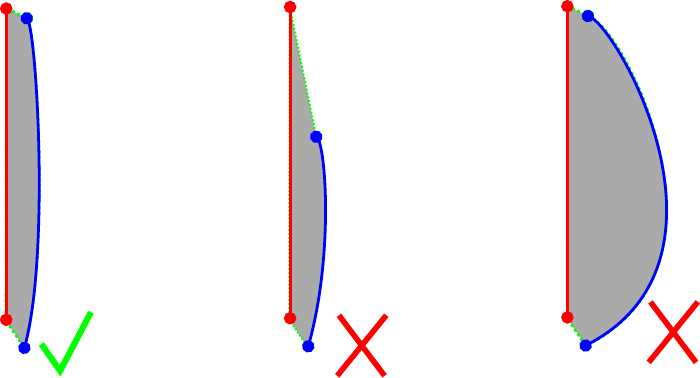}
		\caption{This figuration visualizes the evaluation error. The red line is the ground truth, and the blue line is the prediction result to be evaluated. The second example mismatched because the distance between endpoints is too far, the third example mismatched because the area(gray part) between two lines are too large.}
		\label{fig:eval}
	\end{figure}

	The metric in a way is just a representation of how close two lines are, as is shown in Fig~\ref{fig:eval}. In our experiments, \texttt{disThresh} is set to 40px due to the high-resolution of images and \texttt{areaThresh} is set to \texttt{disThresh} times the length of \texttt{gtLine}.
	
\begin{figure}
		\centering
		\if\smallpics1
		\includegraphics[height=2.5cm]{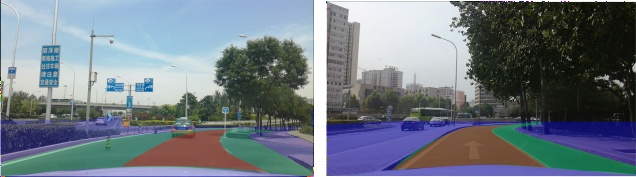}
		\else
		\includegraphics[height=2.5cm]{figs/lane_road.png}
		\fi
		\caption{Differences between ego-lane(red), ego-road(green and red), and all roads(blue, green and red).}
		\label{fig:ego}
	\end{figure}

	We also provided two different evaluation settings: ego-road (i.e., the road that the vehicle is currently driving on) evaluation and all roads evaluation. Ego-road evaluation only considers lane markings on the ego-road, while all roads evaluation considers all lane markings. Differences among ego-lane, ego-road, and all roads are shown in Fig~\ref{fig:ego}.
	
	LineNet was compared with three state-of-the-art methods(Mask R-CNN\cite{He2017MaskR}, MLD-CRF\cite{Hur2013MultilaneDI}, and SCNN\cite{pan2018SCNN}) on the two settings, {\em i.e.}, ego-road evaluation and all roads evaluation. We did not  compare the method with VPGNet\cite{Lee2017VPGNetVP} because the vanishing points are not annotated in our dataset. Results are shown in Table~\ref{table:ttlane}. Our method achieved the best performance on all evaluation metrics.
	
	\setlength{\tabcolsep}{4pt}
	\begin{table}
		\begin{center}
			\caption{Comparison of comprehensive lane detection.}
			\label{table:ttlane}
			\begin{tabular}{l|ccc|ccc}
				\hline
				\multirow{2}{4em}{Method}  & \multicolumn{3}{|c|}{ego-road} & \multicolumn{3}{|c}{all roads} \\
				\cline{2-7}
				& F1 & PRE & REC & F1 & PRE & REC \\
				\hline
				Mask R-CNN\cite{He2017MaskR} & 0.568 & 0.580 & 0.556 & 0.521 & 0.538 & 0.506  \\
				MLD-CRF\cite{Hur2013MultilaneDI}  & 0.206 & 0.260 & 0.171 & 0.193 & 0.267 & 0.150\\
				SCNN\cite{pan2018SCNN} & 0.608 & 0.594 & 0.623  & 0.560 & 0.537 & 0.585 \\
				Ours  & \textbf{0.708} & \textbf{0.651} & \textbf{0.778} & \textbf{0.663} & \textbf{0.613}  & \textbf{0.722} \\
				\hline
			\end{tabular}
		\end{center}
	\end{table}
	\setlength{\tabcolsep}{1.4pt}
	
\begin{figure}
		\centering
		\includegraphics[height=9cm]{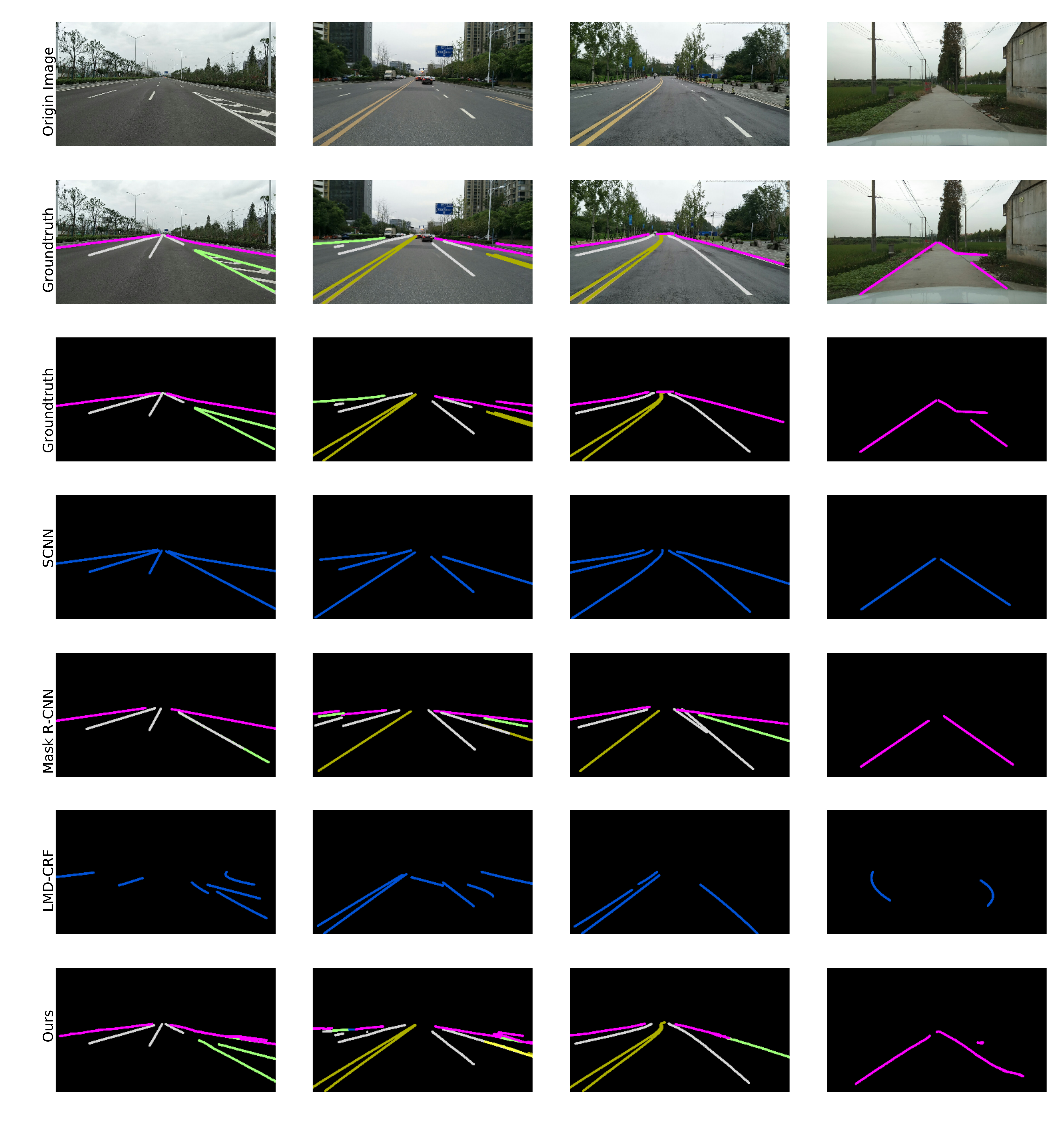}
		\caption{Visual compression on TTLane testing set. LineNet performs better in double lines and line direction on the remote side.}
		\label{fig:compare}
	\end{figure}

	To intuitively show the advantages of LineNet, here in Fig~\ref{fig:compare} some prediction results on the testing set are shown. These results were compared to Mask R-CNN\cite{He2017MaskR} and MLD-CRF\cite{Hur2013MultilaneDI} and SCNN\cite{pan2018SCNN}. As is shown in column 1-3, due to the benefit of zoom module, LineNet can predict double lines as two separate lines, while Mask-RCNN and SCNN cannot. In column 3, we can see clearly that LineNet does well on the curve on the remote side (yellow lines), while Mask R-CNN and MLD-CRF just hold the line straight. In column 4, LineNet can still work well in the rural environment, however, MLD-CRF breaks down. We can conclude from the experiment that mask-based lane detection, like Mask R-CNN and SCNN, cannot handle details well, like double line detection very well. Methods like MLD-CRF that don't need training can hardly perform well in complex scenes.SCNN\cite{pan2018SCNN} and MLD-CRF\cite{Hur2013MultilaneDI} do not produce classes information, so the classes with all methods were evaluated except these two. By contrast, LineNet was robust and performed well in both situations.
	
	\subsection{HD Map Modeling}
	\label{sec:hdmap_exp}

	\begin{figure*}
		\centering
		% ramp 匝道
		\subfigure[This sub figuration showed road modeling results in lane changed. As you can see, our method can model the ramp and multiple lanes.]{
			\if\smallpics1
			\fbox{\includegraphics[width=12.0cm]{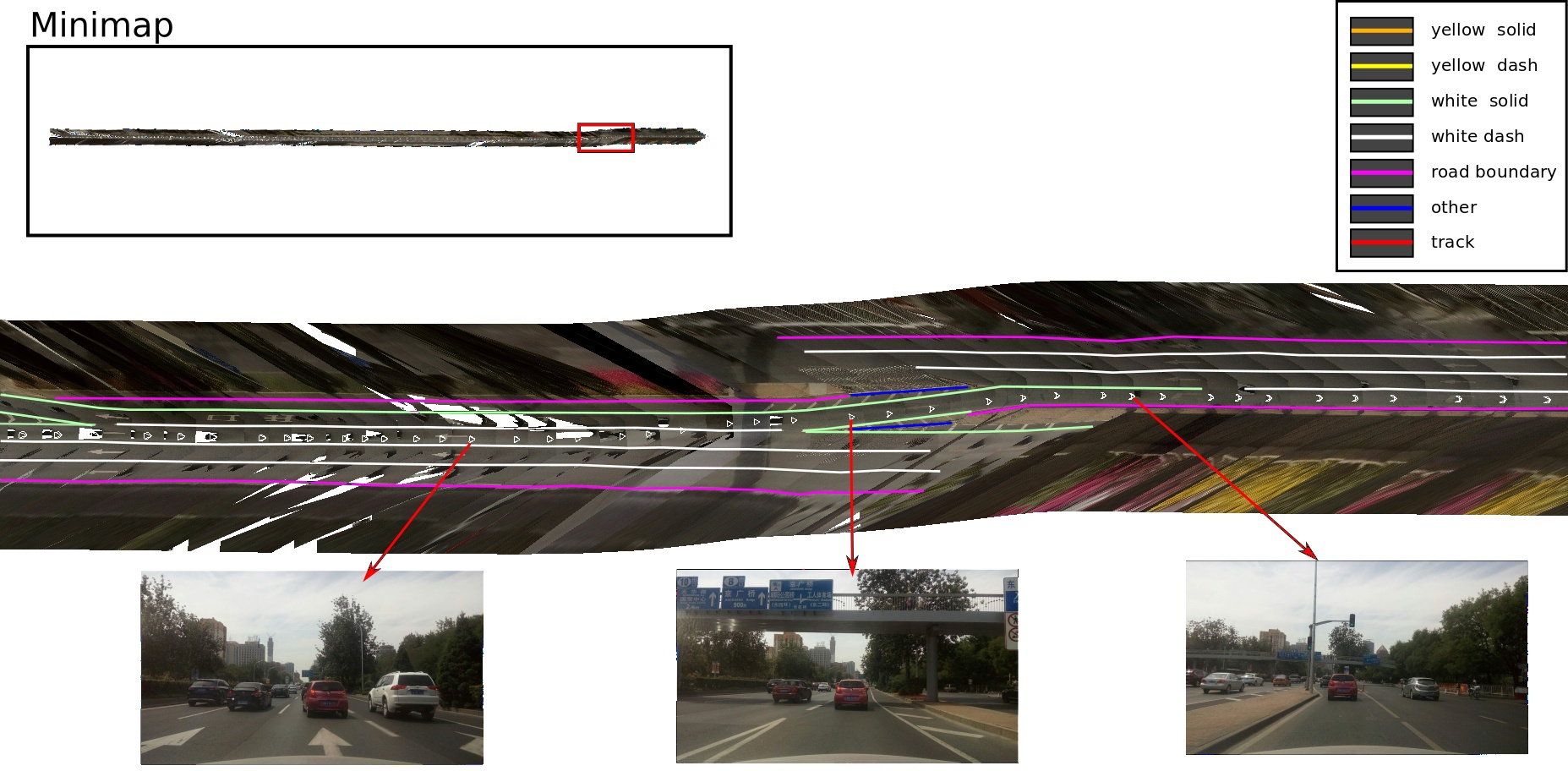}}
			\else
			\fbox{\includegraphics[width=12.0cm]{figs/sfm/ff/ff1.png}}
			\fi
		}
		\subfigure[This sub figuration showed road modeling results in turning. SfM is stabled when turning, and the road modeling results were still clean and well matched.]{
			\if\smallpics1
			\fbox{\includegraphics[width=12.0cm]{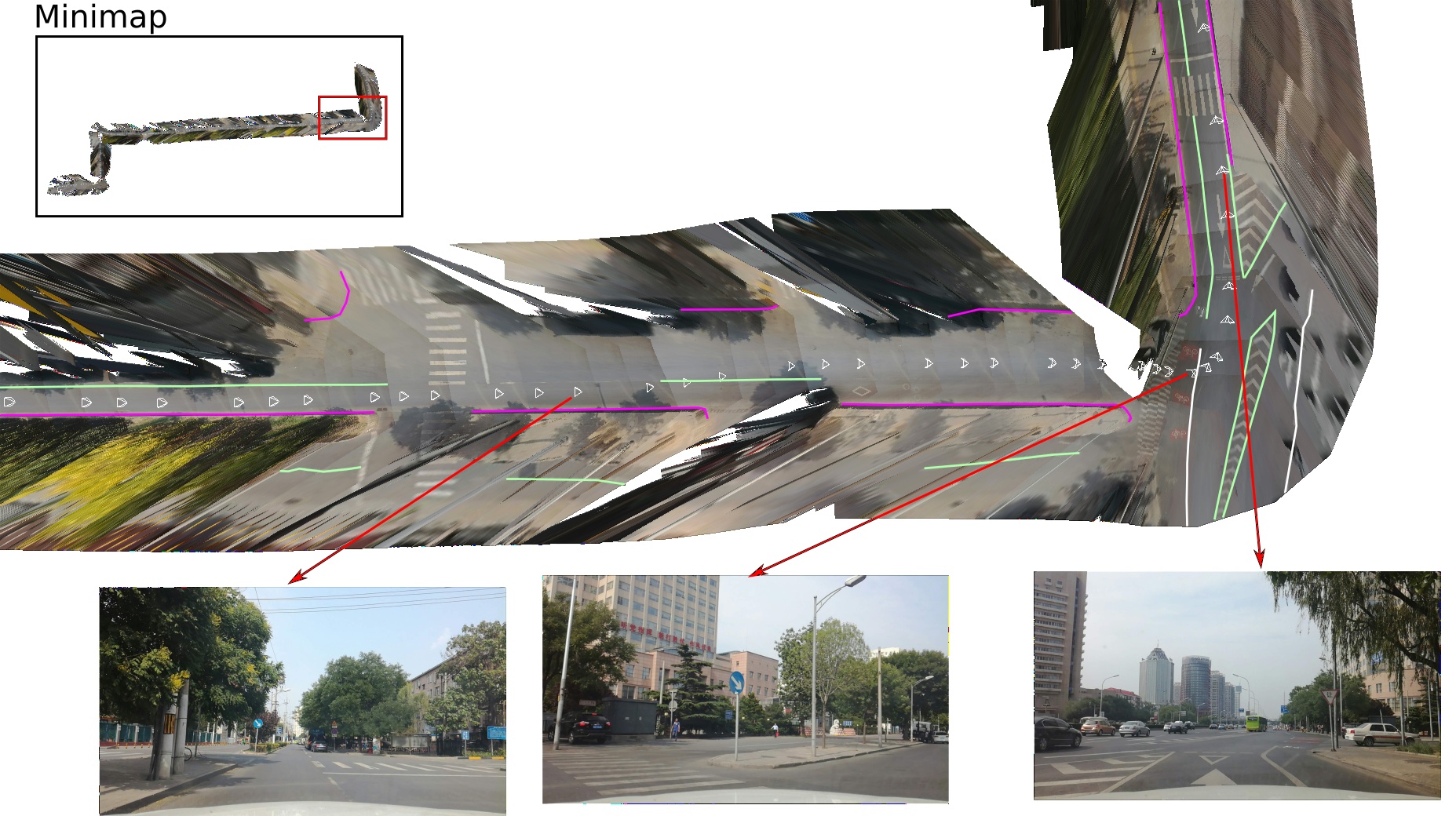}}
			\else
			\fbox{\includegraphics[width=12.0cm]{figs/sfm/ff/ff2.png}}
			\fi
		}
		\subfigure[This sub figuration showed road modeling results in crossroads. This data collected from three different transportation. In the zooming area, two drivers passed crossroad in different directions. SfM can stitch this two road together and yield self-conform results.]{
			\if\smallpics1
			\fbox{\includegraphics[width=12.0cm]{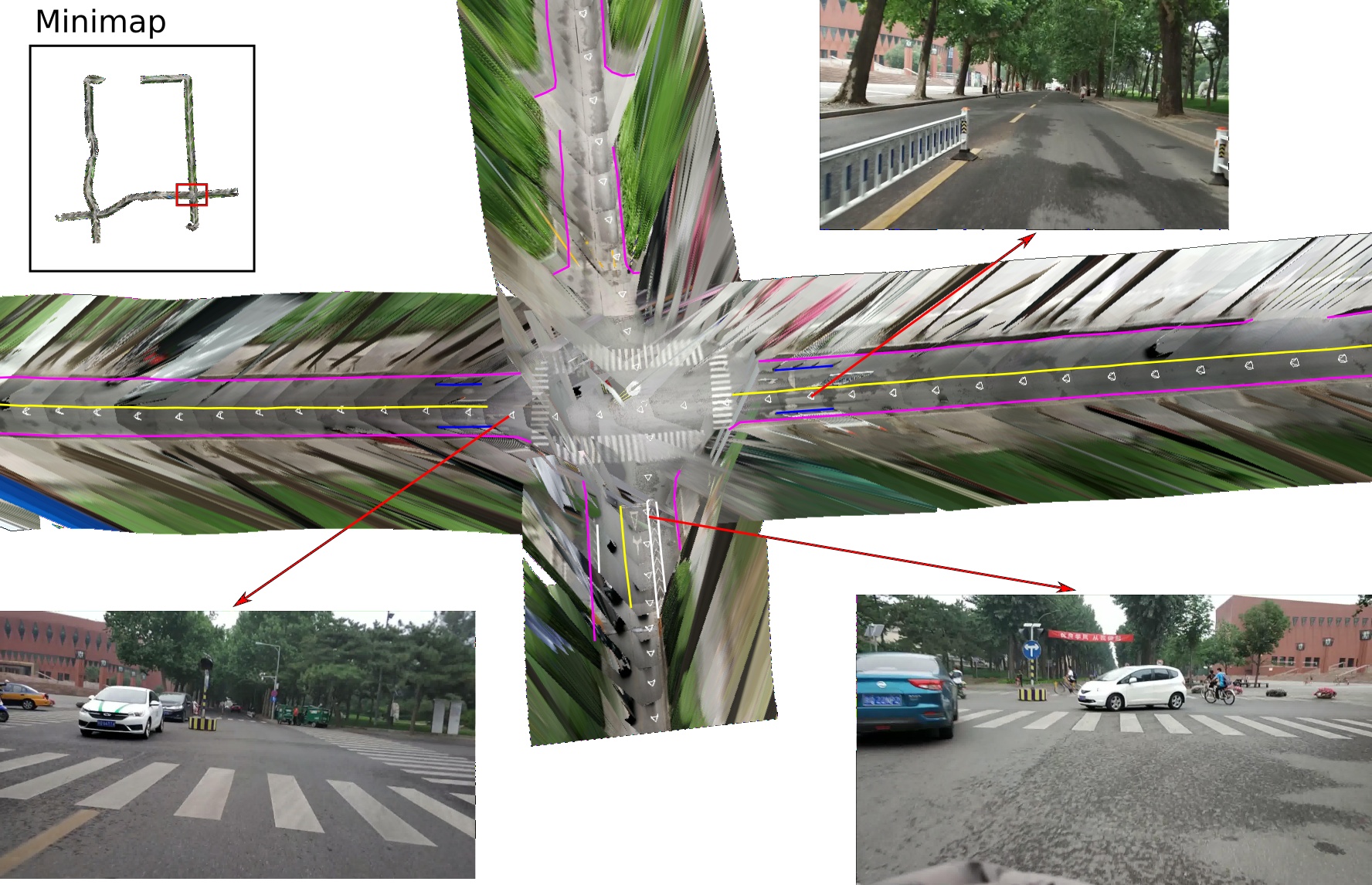}}
			\else
			\fbox{\includegraphics[width=12.0cm]{figs/sfm/ff/ff3.png}}
			\fi
		}
		\caption{This three sub figurations showed how our road modeling algorithm works on three common road conditions. Each sub figurations contained a mini-map and a zooming area, the red box in the mini-map indicated the location of the zooming area. In the zooming area, the road modeling results were shown with colored lines. Each white triangle represented a shot position. The interval between two shots is about 5-20 meters, because of the limitation of bandwidth. Note: We did not use any additional data(such as satellite image) except images and GPS, the ground surface images were stitch from crowdsourced data.}
		\label{fig:merge_lane}
	\end{figure*}
	
	To test our pipeline for HD maps modeling, we ran a small scale experiment and successfully constructed HD maps with inaccurate data. This pipeline consists of OpenSfM\footnote{https://github.com/mapillary/OpenSfM}, LineNet and some post-processing. In this experiment, images and GPS information were collected with different transportations (cars, bicycles, electric bikes) in three road conditions (straight, turning, and crossroads). These transportations were driven by three riders. These riders covered different sections of the road and overlapped with each other. Images were collected at intervals of 5-20 meters, combined with inaccurate GPS information (with errors about 5 meters). HD maps were successfully constructed with our pipeline. It was found that the SfM is well connected to the overlapping parts and the post-processing converted detection results to smooth and reliable lines. Ground truth of these roads was also collected. Compared with the ground truth, our crowdsourced method can reduce the error to about 31.3 cm. It is worth mentioning that although the GPS error of mobile devices is large, the method can still reduce the error to a small amount.
	
	In this section, two kinds of error measurement are used: ground surface parameters reconstruction error, that reflects the error of using SfM to rebuild the ground surface, and lane error, that reflects the lane offset with the ground truth.
	%在这一节，我们需要做了两种误差测量，Road Surface reconstruction error and Lane error，Road Surface reconstruction error 反应了利用SfM重建路面的误差，而Lane error反映了车道线偏移量
	
	\subsubsection{Ground Surface Parameters Reconstruction}
	
	%我们对SfM得到的点云信息进行segmentation，得到所有地面点云，然后将所有地面点云拟合到一个曲面上，将这个曲面作为地面
	\begin{figure}
		\centering
		
		\if\smallpics1
		\includegraphics[width=6.0cm]{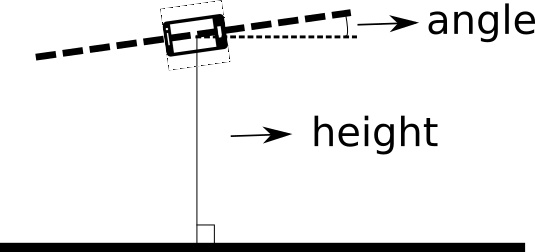}
		\else
		\includegraphics[width=6.0cm]{figs/ground_param.png}
		\fi
		\caption{Ground Surface Parameters consist of angle and height. Angle parameter is the angle between the horizontal line of the mobile phone and the ground. Height parameter is the distance between the mobile phone and the ground.}
		\label{fig:gsp}
	\end{figure}
	Compared with the original GPS information, using SfM can reconstruct more accurate camera movement. But the road surface is still unknown. To construct the road surface, we segmented the point cloud obtained by SfM, and fit all the ground point clouds to a surface with ground surface parameters. Then the surface was used as the ground. Ground Surface Parameters(GSPs) consist of angle and height(showed in Fig~\ref{fig:gsp}), we solved the GSPs for each shot position, Table~\ref{table:gsp} showed our GSPs reconstruction error in three different road conditions. These three road conditions were shown in Fig~\ref{fig:merge_lane}.
	
	\begin{table}
		
		\begin{center}
			\caption{GSP reconstruction error}
			\label{table:gsp}
			\begin{tabular}{l|c|c|c}
				\hline
				Data & Shots Number & Average Height Error & Average Rotation Error   \\
				\hline
				straight & 173 & 0.15 meters & 5 degrees \\
				turning & 342 & 0.11 meters & 8 degrees \\
				crossroad & 1476 & 0.23 meters & 9 degrees \\
				\hline
			\end{tabular}
		\end{center}
	\end{table}
	
	\subsubsection{Lane Merging}
	
	The lanes that LineNet detected were still fragmented. There is no correlation between shots. To model entire lanes, these detection results were projected into the road surface, and they were merged by the same method we mentioned in Section~\ref{sec:post}. Fig~\ref{fig:merge_lane} shows the lanes were merged in three different road conditions. To verify our lane merging results, we evaluated our result to the ground-truth which collected from laser sensors. Evaluate method is the same with Section~\ref{sec:cld}. The average evaluation error is 31.3 cm. This is a significant improvement because the GPS error of our mobile devices is about 5 meters.
	
	The HD map modeling experiment is still small due to our limited resources. We plan to model a city-scale HD maps and release it into a mobile phone application for everyone. 
	
	\subsection{Ablation Study}
	
	\subsubsection{Training with Occlusion Segments Improves Accuracy}
	\label{exp_occl}
	Lane marks can be blocked by moving vehicles and thus be incomplete. We call the blocked parts of the mark ``occlusion segments". Our TTLane Dataset annotates occlusion segments. To inspect the effect of occlusion segments. LineNet was trained under two conditions. one is to treat occluded segments as normal lane markings, called ``training without occlusion segments". Another is to treat them as a special type of lines, called ``training with occlusion segments". Results are shown in Table~\ref{table:occlusion}. The experiment showed that training with occlusion segments performed better. 
	\setlength{\tabcolsep}{4pt}
	\begin{table}
		\begin{center}
			\caption{Comparison on Ablation Study for Occlusion Segments}
			\label{table:occlusion}
			\begin{tabular}{llll}
				\hline\noalign{\smallskip}
				Strategy & F1-score & Precision  & Recall \\
				\noalign{\smallskip}
				\hline
				\noalign{\smallskip}
				without occlusion segments & 0.637 & 0.553 & 0.753 \\
				with occlusion segments  & \textbf{0.708} & \textbf{0.651} & \textbf{0.778} \\
				\hline
			\end{tabular}
		\end{center}
	\end{table}
	\setlength{\tabcolsep}{1.4pt}

	\subsubsection{Zoom Module Improves Precision}
	
	\setlength{\tabcolsep}{4pt}
	\begin{table}
		\begin{center}
			\caption{Comparison on Ablation Study for Occlusion Segments}
			\label{table:zoom_module_ab}
			\begin{tabular}{llll}
				\hline\noalign{\smallskip}
				Strategy & F1-score & Precision  & Recall \\
				\noalign{\smallskip}
				\hline
				\noalign{\smallskip}
				without zoom modules & 0.634 & 0.525 & \textbf{0.801} \\
				with zoom modules  & \textbf{0.708} & \textbf{0.651} & 0.778 \\
				\hline
			\end{tabular}
		\end{center}
	\end{table}
	\setlength{\tabcolsep}{1.4pt}
	
	%在这个试验中，我们取消了Zoom module的特征融合过程，这会导致tumb net的全局特征无法传递给hcrop net，我们发现，特征融合过程可以显著提升precision，但是recall会有轻微的下降，F1-score提升， 结果显示在了表中。
	
	In this experiment, we canceled the feature fusion process of the zoom module, resulting in that the global features of the thumbnail net cannot be transmitted to the high-resolution crops net. We found that the feature fusion process significantly improved precision, but the recall had a slight decrease. The F1-score was improved with the results appearing in the Table~\ref{table:zoom_module_ab}.
	
	\subsubsection{The Influences of each branch in LP layer}
	
	%在这个试验中，我们分离了不同的分支进行实验，可以发现，position分支在我们使用阈值为40像素的评测时，使用position分支轻微改进了f1score，但当我们将阈值设置为10像素时，差距被拉大了，使用position分支对效果有显著的改善。
	In this experiment, we separated the different branches, including position branch, direction branch, distance branch, and confidence branch. The position branch produces more accurate results. When we use the \texttt{disThres} of 40 pixels, the F1-score was only slightly improved by the position branch. But when the \texttt{disThres} was set to 10 pixels, using the position branch significantly improved the effect(Table~\ref{table:bc_ab}).
	
	%direction分支和distance分支都对最终结果影响较小，但我们观察训练过程可以发现，这两个分支可以让网络收敛速度提升，10000步的时候，误差比不使用这两个分支小大约10% 。
	Both the direction branch and the distance branch improve the convergence speed during the training process. In 10,000 steps, the training loss is about 10\% smaller when the two branches were used. And using this two branches does not affect the F1-score of final result.
	
	% 当我们去除掉confidence分支时，zoom in process也同时关闭了，因为zoom in process依赖与confidence分支的结果，所以网络只会看到第一个尺度的图片，效果大幅下降。
	
	The confidence branch is necessary for increasing the F1-score. 
	When we removed the confidence branch, the zooming process also turned off because it depends on the result of the confidence branch, so the network will only see the image in the first stage. And the F1-score was greatly reduced.
	
	\setlength{\tabcolsep}{4pt}
	\begin{table}
		\begin{center}
			\caption{Comparison on Ablation Study for each branches}
			\label{table:bc_ab}
			\begin{tabular}{lllll}
				\hline\noalign{\smallskip}
				Strategy & F1-score & Precision  & Recall & F1-score(10px) \\
				\noalign{\smallskip}
				\hline
				\noalign{\smallskip}
				without position branch & 0.697 & 0.643 & 0.762  & 0.627 \\
				without direction branch  & 0.703 & 0.647 & 0.771  & 0.673 \\
				without distance branch  & 0.696 & 0.640 & 0.763  & 0.670 \\
				without confidence branch  & 0.615 & 0.554 & 0.693  & 0.531 \\
				with all branches  & \textbf{0.708} & \textbf{0.651} & \textbf{0.778} & \textbf{0.682} \\
				\hline
			\end{tabular}
		\end{center}
	\end{table}
	\setlength{\tabcolsep}{1.4pt}

	\section{Conclusion}
	In this paper, we propose LineNet and the TTLane dataset, both of which help us capture the lane and road boundary for crowdsourced HD maps modeling. For the first time, we propose a pipeline to model HD Maps with crowdsourced data our method achieves great precision even with inaccurate data. The method also shows the ability to model HD maps accurately without special equipment. Therefore our method can enlarge the coverage area of HD maps efficiently.

	% if have a single appendix:
	%\appendix[Proof of the Zonklar Equations]
	% or
	%\appendix  % for no appendix heading
	% do not use \section anymore after \appendix, only \section*
	% is possibly needed
	
	% use appendices with more than one appendix
	% then use \section to start each appendix
	% you must declare a \section before using any
	% \subsection or using \label (\appendices by itself
	% starts a section numbered zero.)
	%
	\appendices
	\section*{Acknowledgment}
	
	This work is supported by Tencent Group Limited. We
	would like to thank Luoxing Xiong, Zhe Zhu, and Mingming Cheng for giving a lot of valuable advices in this work.
	
	\iffalse
	\section{}
	Appendix one text goes here.
	
	% you can choose not to have a title for an appendix
	% if you want by leaving the argument blank
	\section{}
	Appendix two text goes here.

	% use section* for acknowledgment
	\fi	
	
	% Can use something like this to put references on a page
	% by themselves when using endfloat and the captionsoff option.
	\ifCLASSOPTIONcaptionsoff
	\newpage
	\fi

\end{document}